# Long Text Generation via Adversarial Training with Leaked Information

**Jiaxian Guo**[†], **Sidi Lu**[†], **Han Cai**[†], **Weinan Zhang**[†*], **Yong Yu**[†], **Jun Wang**[‡]

[†]Shanghai Jiao Tong University, [‡]University College London

{jiaxian,steve_lu,hcai,wnzhang,yyu}@apex.sjtu.edu.cn, j.wang@cs.ucl.ac.uk

## Abstract

Automatically generating coherent and semantically meaningful text has many applications in machine translation, dialogue systems, image captioning, etc. Recently, by combining with policy gradient, Generative Adversarial Nets (GAN) that use a discriminative model to guide the training of the generative model as a reinforcement learning policy has shown promising results in text generation. However, the scalar guiding signal is only available after the entire text has been generated and lacks intermediate information about text structure during the generative process. As such, it limits its success when the length of the generated text samples is long (more than 20 words). In this paper, we propose a new framework, called LeakGAN, to address the problem for long text generation. We allow the discriminative net to leak its own high-level extracted features to the generative net to further help the guidance. The generator incorporates such informative signals into all generation steps through an additional MANAGER module, which takes the extracted features of current generated words and outputs a latent vector to guide the WORKER module for next-word generation. Our extensive experiments on synthetic data and various real-world tasks with Turing test demonstrate that LeakGAN is highly effective in long text generation and also improves the performance in short text generation scenarios. More importantly, without any supervision, LeakGAN would be able to implicitly learn sentence structures only through the interaction between MANAGER and WORKER.

## Introduction

The ability to generate coherent and semantically meaningful text plays a key role in many natural language processing applications such as machine translation (Yang et al. 2017), dialogue generation (Li et al. 2017), and image captioning (Fang et al. 2015). While most previous work focuses on task-specific applications in supervised settings (Bahdanau, Cho, and Bengio 2014; Vinyals et al. 2015), the generic unsupervised text generation, which aims to mimic the distribution over real text from a corpus, has recently drawn much attention (Graves 2013; Yu et al. 2017; Zhang et al. 2017; Hu et al. 2017). A typical approach is to train a recurrent neural network (RNN) to maximize the log-likelihood of each ground-truth word given prior observed words (Graves 2013), which, however, suffers from so-called exposure bias due to the discrepancy between training and inference stage: the model sequentially generates the next word based on previously generated words during inference but itself is trained to generate words given ground-truth words (Huszár 2015). A scheduled sampling approach (Bengio et al. 2015) is proposed to addressed this problem, but is proved to be fundamentally inconsistent (Huszár 2015). Generative Adversarial Nets (GAN) (Goodfellow et al. 2014), which is firstly proposed for continous data (image generation etc.), is then extended to discrete, sequential data to alleviate the above problem and has shown promising results (Yu et al. 2017). Due to the discrete nature of text samples, text generation is modeled as a sequential decision making process, where the state is previously generated words, the action is the next word to be generated, and the generative net $G$ is a stochastic policy that maps current state to a distribution over the action space. After the whole text generation is done, the generated text samples are then fed to the discriminative net $D$, a classifier that is trained to distinguish real and generated text samples, to get reward signals for updating $G$.

Since then, various methods have been proposed in text generation via GAN (Lin et al. 2017; Rajeswar et al. 2017; Che et al. 2017). Nonetheless, the reported results are limited to the cases that the generated text samples are short (say, fewer than 20 words) while more challenging long text generation is hardly studied, which is necessary for practical tasks such as auto-generation of news articles or product descriptions. A main drawback of existing methods to long text generation is that the binary guiding signal from $D$ is sparse as it is only available when the whole text sample is generated. Also, the scalar guiding signal for a whole text is non-informative as it does not necessarily preserve the picture about the intermediate syntactic structure and semantics of the text that is being generated for $G$ to sufficiently learn.

On one hand, to make the guiding signals more informative, discriminator $D$ could potentially provide more guidance beside the final reward value, since $D$ is a trained model, e.g. a convolutional neural network (CNN) (Zhang and LeCun 2015), rather than an unknown black box. With that idea, (Zhang et al. 2017) proposed to train generator $G$ via forcing learned feature representations of real and generated text by $D$ to be matched, instead of directly training $G$ to maximize the reward from $D$ (Yu et al. 2017). Such a method can be effective in short text generation, but the guiding signals are still absent until the end of the text (Zhang et al. 2017).

[*]Correspondence to Weinan Zhang. This work is financially supported by NSFC (61702327) and Shanghai Sailing Program (17YF1428200).

On the other hand, to alleviate the sparsity problem of the guiding signal, the idea of *hierarchy* naturally arises in text generation, since the real text samples are generated following some kinds of hierarchy such as the semantic structure and the part-of-speech (Mauldin 1984). By decomposing the whole generation task into various sub-tasks according to the hierarchical structure, it becomes much easier for the model to learn. Early efforts have been made to incorporate the hierarchy idea in text generation (Dethlefs and Cuayáhuitl 2010; Peng et al. 2017) but all use a predefined sub-task set from domain knowledge, which makes them unable to adapt to arbitrary sequence generation tasks.

In this paper, we propose a new algorithmic framework called LeakGAN to address both the non-informativeness and the sparsity issues. LeakGAN is a new way of providing richer information from the discriminator to the generator by borrowing the recent advances in hierarchical reinforcement learning (Vezhnevets et al. 2017). As illustrated in Figure 1, we specifically introduce a hierarchical generator $G$, which consists of a high-level MANAGER module and a low-level WORKER module. The MANAGER is a long short-term memory network (LSTM) (Hochreiter and Schmidhuber 1997) and serves as a mediator. In each step, it receives generator $D$'s high-level feature representation, e.g., the feature map of the CNN, and uses it to form the guiding goal for the WORKER module in that timestep. As the information from $D$ is internally-maintained and in an adversarial game it is not supposed to provide $G$ with such information. We thus call it a leakage of information from $D$.

Next, given the goal embedding produced by the MANAGER, the WORKER first encodes current generated words with another LSTM, then combines the output of the LSTM and the goal embedding to take a final action at current state. As such, the guiding signals from $D$ are not only available to $G$ at the end in terms of the scalar reward signals, but also available in terms of a goal embedding vector during the generation process to guide $G$ how to get improved.

We conduct extensive experiments based on synthetic and real data. For synthetic data, LeakGAN obtains much lower negative log-likelihood than previous models with sequence length set to 20 and 40. For real data, we use the text in EMNLP2017 WMT News, COCO Image Caption and Chinese Poems as the long, mid-length and short text corpus, respectively. In all those cases, LeakGAN shows significant improvements compared to previous models in terms of BLEU statistics and human Turing test. We further provide a deep investigation on the interaction between MANAGER and WORKER, which indicates LeakGAN implicitly learns sentence structures, such as punctuation, clause structure and long suffix without any supervision.

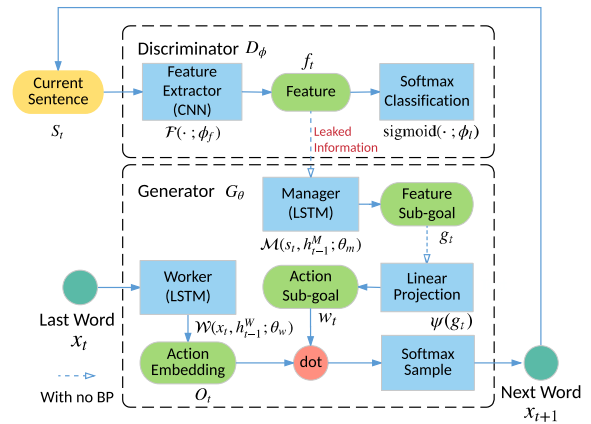

Figure 1: An overview of our LeakGAN text generation framework. While the generator is responsible to generate the next word, the discriminator adversarially judges the generated sentence once it is complete. The chief novelty lies in that, unlike conventional adversarial training, during the process, the discriminator reveals its internal state (feature $f_t$) in order to guide the generator more informatively and frequently. (See Methodology Section for more details.)

## Related Work

Generating text that mimics human's expression has been studied for poem generation (Zhang and Lapata 2014), image captioning (Vinyals et al. 2015), dialogue system (Li et al. 2017) machine translation (Yang et al. 2017). (Graves 2013) proposed a recurrent neural network (RNN) based generative model to use the human-generated text where at each step the model tries to predict the next word given previous real word sequence and is trained in a supervised fashion.

A common difficulty of all supervised generative models is that it is hard to design an appropriate, differentiable, low-bias metric to evaluate the output of the generator, which inspires the adversarial training mechanisms. (Goodfellow et al. 2014) proposed generative adversarial nets (GANs) to generate continuous data like images. GAN introduces a minimax game between a generative model and a discriminative model, where the discriminator can be viewed as the dynamically-updated evaluation metric to guide the tuning of the generated data. To apply GANs to text generation, (Yu et al. 2017) proposed SeqGAN that models the text generation as a sequential decision making process and trains the generative model with policy gradient methods (Sutton et al. 1999). MaliGAN (Che et al. 2017) modifies the orginal GAN objective and proposes a set of training techniques to reduce the potential variance. To deal with the gradient vanishing problem of GAN, RankGAN (Lin et al. 2017) proposes an alternative solution to this problem by replacing the original binary classifier discriminator with a ranking model by taking a softmax over the expected cosine distances from the generated sequences to the real data. Another problem for the adversarial sequence generation models is that the binary feedback from the discriminator is not sufficiently informative, which requires a huge number of training and generated samples to improve the generator and could result in mode collapse problems. Feature Matching (Zhang et al. 2017) provides a mechanism that matches the latent feature distributions of real and generated sequences via a kernelized discepancy metric to alleviate the weak guidance and mode collapse problems. However, such enhancement only happens when the whole text sample is generated and thus the guiding signal is still sparse during the training.

Reinforcement learning (RL) on the other hand also faces a similar difficulty when reward signals are sparse (Kulkarni et al. 2016). Hierarchical RL is one of the promising techniques for handling the sparse reward issue (Sutton, Precup, and Singh 1999). A typical approach in hierarchical RL is to manually identify the hierarchical structure for the agent by defining several low-level sub-tasks and learning micropolicies for each sub-task while learning a macro-policy for choosing which sub-task to solve. Such methods can be very effective when the hierarchical structure is known a priori using domain knowledge in a given specific task, but fail to flexibly adapt to other tasks. Recently, (Vezhnevets et al. 2017) proposed an end-to-end framework for hierarchical RL where the sub-tasks are not identified manually but implicitly learned by a MANAGER module which takes current state as input and output a goal embedding vector to guide the low-level WORKER module.

In this work, we model the text generation procedure via adversarial training and policy gradient (Yu et al. 2017). To address the sparse reward issue in long text generation, we follow (Vezhnevets et al. 2017) and propose a hierarchy design, i.e. MANAGER and WORKER, for the generator. As the reward function in our case is a discriminative model rather than a black box in (Vezhnevets et al. 2017), the high-level feature extracted by the discriminator given the current generated word sequence is sent to the MANAGER module. As such, the MANAGER module can be also viewed as a spy that *leaks* information from the discriminator to better guide the generator. To our knowledge, this is the first work that considers the information leaking in GAN framework for better training generators and combines hierarchical RL to address long text generation problems.

## Methodology

We formalize the text generation problem as a sequential decision making process (Bachman and Precup 2015). Specifically, at each timestep $t$, the agent takes the previously generated words as its current state, denoted as $s_t = (x_1, \ldots, x_i, \ldots, x_t)$, where $x_i$ represents a word token in the given vocabulary $V$. A $\theta$-parameterized generative net $G_\theta$, which corresponds to a stochastic policy, maps $s_t$ to a distribution over the whole vocabulary, i.e. $G_\theta(\cdot|s_t)$, from which the action $x_{t+1}$, i.e. the next word to select is sampled. We also train a $\phi$-parameterized discriminative model $D_\phi$ that provides a scalar guiding signal $D_\phi(s_T)$ for $G_\theta$ to adjust its parameters when the whole sentence $s_T$ has been generated.

As we discussed previously, although the above adversarial training is principled, the scalar guiding signal becomes relatively less informative when the sentence length $T$ goes larger. To address this, the proposed LeakGAN framework allows discriminator $D_\phi$ to provide additional information, denoted as features $f_t$, of the current sentence $s_t$ (it is internally used for $D_\phi$ itself for discrimination) to generator $G_\theta(\cdot|s_t)$. In LeakGAN, a hierarchical RL architecture is used as a promising mechanism to effectively incorporate such leaked information $f_t$ into the generation procedure of $G_\theta$ (also see Figure 1).

### Leaked Features from $D$ as Guiding Signals

Different from typical model-free RL settings where the reward function is a black box, our adversarial text generation uses $D_\phi$ as a learned reward function. Typically, $D_\phi$ is a neural network and can be decomposed into a feature extractor $\mathcal{F}(\cdot; \phi_f)$ and a final sigmoid classification layer with weight vector $\phi_l$. Mathematically, given input $s$, we have

$$D_\phi(s) = \text{sigmoid}(\phi_l^\top \mathcal{F}(s; \phi_f)) = \text{sigmoid}(\phi_l^\top f), \quad (1)$$

where $\phi = (\phi_f, \phi_l)$ and $\text{sigmoid}(z) = 1/(1 + e^{-z})$. $f = \mathcal{F}(s; \phi_f)$ is the feature vector of $s$ in the last layer of $D_\phi$, which is to be leaked to generator $G_\theta$. As is shown in Eq. (1), for a given $D_\phi$, the reward value for each state $s$ mainly depends on the extracted features $f$. As such, the objective of getting a higher reward from $D_\phi$ is equivalent to finding a higher reward region in this extracted feature space $\mathcal{F}(\boldsymbol{S}; \phi_f) = \{\mathcal{F}(s; \phi_f)\}_{s \in \boldsymbol{S}}$. Specifically, our feature extractor $\mathcal{F}(\cdot; \phi_f)$ in $D_\phi$ is implemented by a CNN (Zhang and LeCun 2015); thus $\mathcal{F}(s; \phi_f)$ outputs the CNN feature map vector as $f$ after its convolution-pooling-activation layer. Other neural network models such as LSTM (Hochreiter and Schmidhuber 1997) can also be used to implement $D_\phi$.

Compared to the scalar signal $D_\phi(s)$, the feature vector $f$ is a much more informative guiding signal for $G_\theta$, since it tells what the position of currently-generated words is in the extracted feature space.

### A Hierarchical Structure of $G$

In each step $t$ during the generation procedure, to utilize the leaked information $f_t$ from $D_\phi$, we follow hierarchical RL (Vezhnevets et al. 2017) to have a hierarchical architecture of $G_\theta$. Specifically, we introduce a MANAGER module, an LSTM that takes the extracted feature vector $f_t$ as its input at each step $t$ and outputs a goal vector $g_t$, which is then fed into the WORKER module to guide the generation of the next word in order to approach the higher reward region in $\mathcal{F}(\boldsymbol{S}; \phi_f)$. Next we will first describe the detailed generator model in LeakGAN and then show how the MANAGER and WORKER are trained with the guiding signals from $D_\phi$.

**Generation Process.** The MANAGER and WORKER modules both start from an all-zero hidden state, denoted as $h_0^M$ and $h_0^W$ respectively. At each step, the MANAGER receives the leaked feature vector $f_t$ from the discriminator $D_\phi$, which is further combined with current hidden state of the MANAGER to produce the goal vector $g_t$ as

$$\hat{g}_t, h_t^M = \mathcal{M}(f_t, h_{t-1}^M; \theta_m), \quad (2)$$
$$g_t = \hat{g}_t / \|\hat{g}_t\|, \quad (3)$$

where $\mathcal{M}(\cdot\,;\theta_m)$ denotes the MANAGER module implemented by an LSTM with parameters $\theta_m$ and $h_t^M$ is the recurrent hidden vector of the LSTM.

To incorporate goals produced by MANAGER, a linear transformation $\psi$ with weight matrix $W_\psi$ is performed on a summation over recent $c$ goals to produce a $k$-dimensional goal embedding vector $w_t$ as

$$w_t = \psi\Big(\sum_{i=1}^{c} g_{t-i}\Big) = W_\psi\Big(\sum_{i=1}^{c} g_{t-i}\Big). \qquad (4)$$

Given the goal embedding vector $w_t$, the WORKER module takes the current word $x_t$ as input and outputs a matrix $O_t$, which is further combined with $w_t$ by matrix product to determine the final action space distribution under current state $s_t$ through a softmax

$$O_t, h_t^W = \mathcal{W}(x_t, h_{t-1}^W; \theta_w), \qquad (5)$$
$$G_\theta(\cdot | s_t) = \mathrm{softmax}(O_t \cdot w_t / \alpha), \qquad (6)$$

where $\mathcal{W}(\cdot\,;\theta_w)$ denotes the WORKER module, i.e. an LSTM with $h_t^W$ as its recurrent hidden vector, $O_t$ is a $|V| \times k$ matrix that represents the current vector for all words, thus $O_t \cdot w_t$ yields the calculated logits for all words, and $\alpha$ is the temperature parameter to control the generation entropy.

## Training of $G$

Notice that the above procedure is fully differentiable. One can train $G_\theta$ in an end-to-end manner using a policy gradient algorithm such as REINFORCE (Williams 1992). In LeakGAN, we would hope the MANAGER module to capture some meaningful patterns. Thus, we follow (Vezhnevets et al. 2017) and train the MANAGER and WORKER modules separately, where the MANAGER is trained to predict advantageous directions in the discriminative feature space and the WORKER is intrinsically rewarded to follow such directions. Similar to (Vezhnevets et al. 2017), the gradient of the MANAGER module is defined as

$$\nabla_{\theta_m}^{\mathrm{adv}} g_t = -Q_{\mathcal{F}}(s_t, g_t) \nabla_{\theta_m} d_{\cos}\Big(f_{t+c} - f_t, g_t(\theta_m)\Big), \quad (7)$$

where $Q_{\mathcal{F}}(s_t, g_t) = Q(\mathcal{F}(s_t), g_t) = Q(f_t, g_t) = \mathbb{E}[r_t]$ is the expected reward under the current policy which can be approximately estimated via Monte Carlo search (Sutton et al. 2000; Yu et al. 2017). $d_{\cos}$ represents the cosine similarity between the change of feature representation after $c$-step transitions, i.e. $f_{t+c} - f_t$, and the goal vector $g_t(\theta_m)$[1] produced by MANAGER as in Eq. (2). Intuitively, the loss function is to force the goal vector to match the transition in the feature space while achieving high reward. At the same time, the WORKER is trained to maximize the reward using the REINFORCE algorithm (Williams 1992) as is done in (Yu et al. 2017),

$$\nabla_{\theta_w} \mathbb{E}_{s_{t-1} \sim G}\Big[\sum_{x_t} r_t^I \mathcal{W}(x_t | s_{t-1}; \theta_w)\Big]$$
$$= \mathbb{E}_{s_{t-1} \sim G, x_t \sim \mathcal{W}(x_t | s_{t-1})}[r_t^I \nabla_{\theta_w} \log \mathcal{W}(x_t | s_{t-1}; \theta_w)], \quad (8)$$

---
[1] We use $g_t(\theta_m)$ to explicitly show $g_t$ is parameterized by $\theta_m$.

which can be approximated by sampling the state $s_{t-1}$ and the action $x_t$ taken by WORKER. As the WORKER is encouraged to follow the directions produced by the MANAGER, following (Vezhnevets et al. 2017), the intrinsic reward for the WORKER is defined as

$$r_t^I = \frac{1}{c} \sum_{i=1}^{c} d_{\cos}\Big(f_t - f_{t-i}, g_{t-i}\Big). \qquad (9)$$

In practice, before the adversarial training, we need to pre-train $G_\theta$. To be consistent, in the pre-train stage, we also use the separate training scheme, where the gradient of MANAGER is

$$\nabla_{\theta_m}^{\mathrm{pre}} g_t = -\nabla_{\theta_m} d_{\cos}\Big(\hat{f}_{t+c} - \hat{f}_t, g_t(\theta_m)\Big), \qquad (10)$$

where $\hat{f}_t = \mathcal{F}(\hat{s}_t)$, $\hat{s}_t$ and $\hat{s}_{t+c}$ are states of real text, and the state-action value $Q_{\mathcal{F}}(s_t, g_t)$ in Eq. (7) is set as 1 here since the data instances used in pre-training are all real sentences. As such, the MANAGER is trained to mimic the transition of real text samples in the feature space. While the WORKER is trained via maximum likelihood estimation (MLE).

In the training process, the generator $G_\theta$ and discriminator $D_\phi$ are alternatively trained. In the generator, the MANAGER $\mathcal{M}(\cdot\,;\theta_m)$ and WORKER $\mathcal{W}(\cdot\,;\theta_w)$ (including $\psi$ and softmax) are alternatively trained while fixing the other. The details of the training procedure are attached in the supplementary material[2].

## Training Techniques

**Bootstrapped Rescaled Activation.** During the adversarial training of SeqGAN (Yu et al. 2017), severe gradient vanishing occurs when $D$ is much stronger than $G$, i.e. the reward is too small value to update the parameters and thus need be rescaled before being fed into $G$. Inspired by ranking idea from RankGAN (Lin et al. 2017), we propose a simple, time-efficient, rank-based method to rescale the rewards, named as *bootstrapped rescaled activation*. For a mini-batch with $B$ sequences, after the rollout of the generative model, the reward matrix is denoted as $R_{B \times T}$. For each timestep $t$, we rescale the $t$-th column vector $R^t$ via

$$R_i^t = \sigma\Big(\delta \cdot \Big(0.5 - \frac{\mathrm{rank}(i)}{B}\Big)\Big), \qquad (11)$$

where $\mathrm{rank}(i)$ denotes the $i$-th element's high-to-low ranking in this column vector. $\delta$ is a hyperparameter that controls the smoothness of the rescale activation. $\sigma(\cdot)$ is an activation function that re-projects the equidifferent scoring based on ranking to a more effective distribution. In our experiment, for example, the model adopts hyperparameter $\delta = 12.0$ and the sigmoid function as $\sigma(\cdot)$.

There are two main advantages of the bootstrapped rescaled activation. First, after this transformation, the expectation and variance of the reward in each mini-batch are constant. In this case, the rescale activation serves as a value stabilizer that is helpful for algorithms that are sensitive in numerical variance. Second, as all ranking methods do, it prevents the gradient vanishing problem, which accelerates the model convergence.

---
[2] https://arxiv.org/abs/1709.08624

**Interleaved Training.** In traditional generative adversarial models, mode collapse is a common problem. Here we propose a training scheme called interleaved training to alleviate such a problem. As its name is, we adopt an interleaving of supervised training (i.e. MLE) and adversarial training (i.e. GAN) instead of full GAN after the pre-training. For example, we perform one epoch of supervised learning for $G$ after 15 epochs of adversarial training. An explanation of why this scheme works is that blending these two trainings would help GAN get rid of some bad local minimums and alleviate mode collapse. Another justification is that the inserted supervised learning performs an implicit regularization on the generative model to prevent it from going too far away from the MLE solution.

**Temperature Control.** The Boltzmann temperature $\alpha$ in Eq. (6) is a factor that could be used to balance the exploration and exploitation for reinforcement learning problems. Here we select a higher temperature when we are training the model and a lower temperature when we adopt the model to generate samples.

# Experiment

The experiment consists of three parts: synthetic data experiments, experiments in real-world scenarios and some explanation study. The repeatable experiment code is published for further research[3].

## Training Settings

**Synthetic Oracle.** For the synthetic data experiments, simlar to (Yu et al. 2017), we first initialize the parameters of an LSTM following the normal distribution $\mathcal{N}(0, 1)$ as the oracle describing the real data distribution $G_{\text{oracle}}(x_t|x_1, \ldots, x_{t-1})$. We use it to generate 10,000 sequences of length 20 and 40 respectively as the training set $\mathcal{S}$ for the generative models.

**GAN Setting.** For the discriminator, we choose the CNN architecture (Zhang and LeCun 2015) as the feature extractor and the binary classifier. Note that one could design specific structure for different tasks to refine the CNN performance. For the synthetic data experiment, the CNN kernel size ranges from 1 to $T$. The number of each kernel is between 100 and 200. In this case, the feature of text is a 1,720 dimensional vector. Dropout (Srivastava et al. 2014) with the keep rate 0.75 and L2 regularization are performed to avoid overfitting. For the generator, we adopt LSTM (Hochreiter and Schmidhuber 1997) as the architectures of MANAGER and WORKER to capture the sequence context information. The MANAGER produces the 16-dimensional goal embedding feature vector $w_t$ using the feature map extracted by CNN. The goal duration time $c$ is a hyperparameter set as 4 after some preliminary experiments.

**Compared Models.** For most parts of our experiment, three baseline models are mainly compared with LeakGAN, namely an MLE trained LSTM, SeqGAN (Yu et al. 2017) and RankGAN (Zhang et al. 2017). We also compare model variants, such as SeqGAN with bootstrapped rescaled activation, and include the real data to be referred as the performance upperbound.

[3] https://github.com/CR-Gjx/LeakGAN.

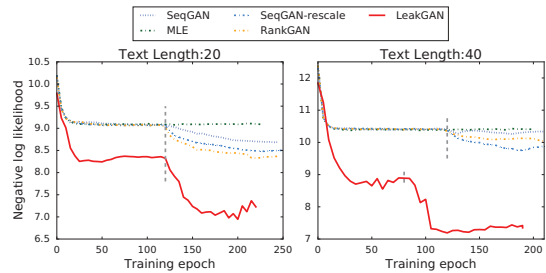

Figure 2: The illustration of training curve.

Table 1: The over NLL performance on synthetic data.

| Length | MLE | SeqGAN | RankGAN | LeakGAN | Real | $p$-value |
|---|---|---|---|---|---|---|
| 20 | 9.038 | 8.736 | 8.247 | **7.038** | 5.750 | $< 10^{-6}$ |
| 40 | 10.411 | 10.310 | 9.958 | **7.191** | 4.071 | $< 10^{-6}$ |

Table 2: BLEU scores performance on EMNLP2017 WMT.

| Method | SeqGAN | RankGAN | LeakGAN | $p$-value |
|---|---|---|---|---|
| BLEU-2 | 0.8590 | 0.778 | **0.956** | $< 10^{-6}$ |
| BLEU-3 | 0.6015 | 0.478 | **0.819** | $< 10^{-6}$ |
| BLEU-4 | 0.4541 | 0.411 | **0.627** | $< 10^{-6}$ |
| BLEU-5 | 0.4498 | 0.463 | **0.498** | $< 10^{-6}$ |

**Evaluation Metrics.** Negative log-likehood (NLL) is used for synthetic data experiment since there is the oracle data distribution available for evaluation. For real-world data experiments, BLEU statistics (Papineni et al. 2002) and human rating scores in the Turing test are reported. We further perform a t-test for the improvement of LeakGAN over the second highest performance and report the $p$-value.

## Synthetic Data Experiments

We run the synthetic data experiment with the text-length set as 20 and 40 respectively.

The training curves are depicted in Figure 2 and the overall NLL performance is presented in Table 1. One could have two observations from the results. (i) In the pre-training stage, LeakGAN has already shown observable performance superiority compared to other models, which indicates that the proposed hierarchical architecture itself brings improvement over the previous ones. (ii) In the adversarial training stage, LeakGAN shows a better speed of convergence, and the local minimum it explores is significantly better than previous results. The results demonstrate the effectiveness of the information leakage framework and the hierarchical RL architecture for generating both short and long texts.

## Long Text Generation: EMNLP2017 WMT News

We choose the EMNLP2017 WMT[4] Dataset as the long text corpus. Specifically, we pick the News section from the original dataset. The news dataset consists of 646,459 words and 397,726 sentences. We preprocess the data by eliminating the words with frequency lower than 4,050 as well as the sentence containing these low frequency words. Besides, to focus on long sentences, we remove the sentences with length less than 20. After the preprocessing, the news dataset has 5,742 words and 397,726 sentences. Then we randomly sample 200,000 sentences as the training set and another 10,000 sentences as the test set. We use the BLEU-(2 to 5) scores (Papineni et al. 2002) as the evaluation metrics.

[4] http://statmt.org/wmt17/translation-task.html

Table 3: BLEU scores on COCO Image Captions.

| Method | SeqGAN | RankGAN | LeakGAN | $p$-value |
|---|---|---|---|---|
| BLEU-2 | 0.831 | 0.850 | **0.950** | $< 10^{-6}$ |
| BLEU-3 | 0.642 | 0.672 | **0.880** | $< 10^{-6}$ |
| BLEU-4 | 0.521 | 0.557 | **0.778** | $< 10^{-6}$ |
| BLEU-5 | 0.427 | 0.544 | **0.686** | $< 10^{-6}$ |

Table 4: The BLEU performance on Chinese Poems.

| Method | SeqGAN | RankGAN | LeakGAN |
|---|---|---|---|
| BLEU-2 | 0.738 | 0.812 | **0.881** |
| $p$-value | $< 10^{-6}$ | $< 10^{-6}$ | - |

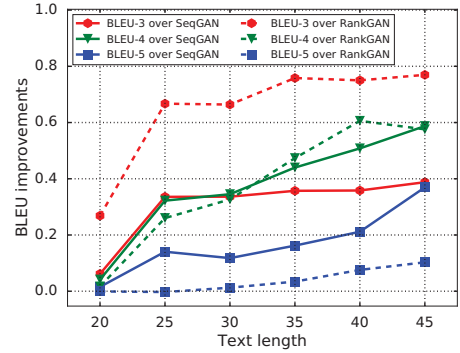

Figure 3: The illustration of BLEU improvement change along with the generated text length on WMT News.

Table 5: Turing test results for in real-world experiments.

| Dataset | SeqGAN | LeakGAN | Ground Truth | $p$-value |
|---|---|---|---|---|
| WMT News | 0.236 | **0.554** | 0.651 | $< 10^{-6}$ |
| COCO | 0.405 | **0.574** | 0.675 | $< 10^{-6}$ |

The results are provided in Table 2. In all measured metrics, LeakGAN shows significant performance gain compared to baseline models. The consistently higher BLEU scores indicate that the generated sentences of LeakGAN are of high quality in local features to mimic the real text.

### Middle Text Generation: COCO Image Captions

Another real dataset we use is the COCO Image Captions Dataset (Chen et al. 2015), a dataset which contains groups of image-description pairs. We take the image captions as the text to generate. Note that the COCO Dataset is not a long text dataset, in which most sentences are of about 10 words. Thus we apply some preprocessing on the dataset. The COCO Image Captions training dataset consists of 20,734 words and 417,126 sentences. We remove the words with frequency lower than 10 as well as the sentence containing them. After the preprocessing, the dataset includes 4,980 words. We randomly sample 80,000 sentences for the training set, and another 5,000 for the test set.

The results BLEU scores are provided in Table 3. The results of the BLEU scores on the COCO dataset indicate that LeakGAN performs significantly better than baseline models in mid-length text generation task.

### Short Text Generation: Chinese Poems

To evaluate the performance of LeakGAN in short text generation, we pick the dataset of Chinese poems which is proposed by (Zhang and Lapata 2014) and most related work such as (Yu et al. 2017; Rajeswar et al. 2017; Lin et al. 2017). The dataset consists of 4-line 5-character poems. Following the above work, we use the BLEU-2 scores as the evaluating metrics.

The experimental results are provided in Table 4. The results on Chinese Poems indicate that LeakGAN successfully handles the short text generation tasks.

### Performance Robustness in Long Text Generation

Long text generation has always been difficult among all text generation problems. The difficulty of the problem is due to many factors, such as LSTM-RNN's failure to capture long-term dependency, discriminator's failure to give those "good but tiny" sequences appropriate penalty. To explicitly evaluate the superiority of LeakGAN in long text generation, here we use the relative performance gain of LeakGAN over SeqGAN (Yu et al. 2017) and RankGAN (Lin et al. 2017).

The results over EMNLP2017 WMT News data are shown in Figure 3. The curves clearly show that LeakGAN yields larger performance gain over the baselines when the generated sentences are longer. This fact supports our claim that LeakGAN is a robust framework for long text.

### Turing Test and Generated Samples

Since BLEU score is a metric focusing on the local text statistics, which may not be sufficient for evaluating text generation quality, we also conduct a Turing test based on questionnaires on the Internet. In the questionnaire, each (machine generated or real) sentence gets +1 score when it is regarded as a real one, and 0 score otherwise. We conduct the test with text generated by the models trained on WMT News and COCO Image Captions. The average score for each algorithm is calculated. In practice, we sample 20 sentences from every method and invite 62 people to participate the test, where everyone should judge the quality of 30 sentences from the compared three methods and thus each sentence is judged by 31 people. For the comparison fairness, the sentences used in the questionnaires are randomly sampled. Table 5 gives the results. The performance on two datasets indicates that the generated sentences of LeakGAN are of higher global consistency and better readability than those of SeqGAN.

A few samples generated by LeakGAN are illustrated in Table 6. More samples and their comparison with those from the baseline models are provided in the supplementary material. These samples are collected for the Turing test questionnaires.

Table 6: Samples from different methods on COCO Image Captions and EMNLP2017 WMT News.

| Datasets | LeakGAN | SeqGAN |
|---|---|---|
| COCO Image Captions | (1) A man sitting in front of a microphone with his dog sitting on his shoulder. (2) A young man is holding a bottle of wine in his hand. | (1) A bathroom with tiled walls and a shower on it. (2) A couple of kids in front of a bathroom that is in a bathroom. |
| EMNLP2017 WMT | (1) The American Medical Association said that the militants had been arrested in connection with the murder of the same incident. (2) This is the first time that the Fed has been able to launch a probe into the country's nuclear program. | (1) "I think you should really really leave for because we hadn't been busy, where it goes to one," he wrote. (2) What you have to stop, if we do that, as late, law enforcement and where schools use a list of aid, it can rise. |

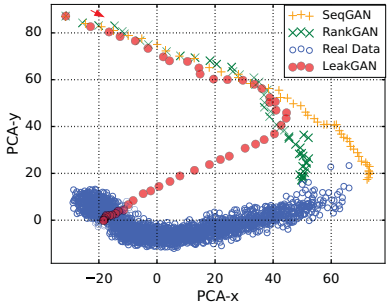

Figure 4: Feature traces during the generation (SeqGAN, RankGAN and LeakGAN) and features of completed real data (all compressed to 2-dim by PCA) on WMT News.

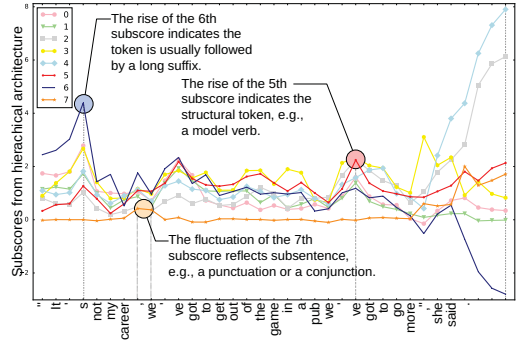

Figure 5: Illustration of WORKER and MANAGER's behaviors during a generation. (Dimension-wise Product of Worker and Manager)

## Model Explanation

**Feature Trace.** To verify that LeakGAN successfully exploits of the leaked message, we visualize the feature vector $f_T$ extracted from the real data by discriminator. Besides, we visualize the feature trace, i.e. the features $f_t$ of prefix $s_t$ during the generation, for LeakGAN, SeqGAN and RankGAN via a 2-D principal component analysis (PCA).

The visualized traces are plotted in Figure 4 and more cases are presented in the supplementary material. As we can see, during the generation process, in LeakGAN, the feature vector gradually approaches the real data feature vector region. However, previous models, i.e. SeqGAN and RankGAN, fail to match the features even when the generation is completed. This indicates that the proposed LeakGAN does finish its design purpose of exploiting the leaked information from $D_\phi$ to better match the feature vector distributions of real data.

**Behaviors of Worker and Manager.** To give more details of how WORKER and MANAGER interact with each other and make use of the leaked information in the generative model, we visualize the interaction vector of the WORKER and MANAGER, i.e., the dimension-wise product of their output ($O_t \cdot w_t$ as in Eq. (6)). Note that to simplify the explanation, here we reduce the signal dimension from 16 to 8. Figure 5 presents an example sentence and more cases are provided in the supplementary material.

From Figure 5, we find some intuitive interpretations of the implicit rules learned by the interaction of WORKER and MANAGER. (i) The 5th dimension stands for current token's divergence from an entity token. If the 5th value is high, the token would most possibly be a structural token, such as a modal verb, an article or a preposition. (ii) The 6th dimension suggests how long the suffix from current step will be. If a peak occurs in the curve, there must be some token that triggers a long suffix. A frequently occurring example is the formal subject. (iii) Although hard to observe, we do find connections of the 7th dimension and the substructure of a sentence. For example, when the start or the end of a subsentence occurs, there is an observable fluctuation in the 7th dimension. This indicates that the token is most likely to be a punctuation or a conjuction.

## Conclusion and Future work

In this paper, we proposed a new algorithmic framework called LeakGAN for generating long text via adversarial training. By leaking the feature extracted by the discriminator as the step-by-step guiding signal to guide the generator better generating long text, LeakGAN addresses the non-informativeness and sparsity problems of the scalar reward signal in previous GAN solutions. In the extensive experiments with synthetic data and real world data including long, mid-length and short text, LeakGAN achieved significant performance improvement over previous solutions, on both BLEU scores and human ratings. Moreover, the analysis of the results shows that LeakGAN yields larger performance gain when the longer sentences are generated. Finally, we also visualize and explain the efficacy of the guiding signals that LeakGAN learns without any supervision.

For future work, we plan to apply LeakGAN in more natural language process applications like dialogue systems and image captioning by providing more task-specific guiding information. Also, enhancing the capacity of the discriminator to check the global consistency of the whole sentence is a promising direction.

# Appendix

**Formulas for Reference**

**Discriminator**

$$f = \mathcal{F}(s; \phi_f), \tag{1}$$

$$D_\phi(s) = \text{sigmoid}(\phi_l \cdot \mathcal{F}(s; \phi_f)) = \text{sigmoid}(\phi_l, f), \tag{2}$$

**MANAGER of Generator**

$$\hat{g}_t, h_t^M = \mathcal{M}(f_t, h_{t-1}^M; \theta_m), \tag{3}$$

$$g_t = \hat{g}_t / \|\hat{g}_t\|, \tag{4}$$

$$w_t = \psi\Big(\sum_{i=1}^{c} g_{t-i}\Big) = W_\psi\Big(\sum_{i=1}^{c} g_{t-i}\Big). \tag{5}$$

$$O_t, h_t^W = \mathcal{W}(x_t, h_{t-1}^W; \theta_w), \tag{6}$$

$$G_\theta(\cdot|s_t) = \text{sigmoid}(O_t \cdot w_t / \alpha), \tag{7}$$

$$Q(f_t, g_t) = \mathbb{E}[r_t], \tag{8}$$

$$\nabla_{\theta_m}^{\text{adv}} g_t = -Q(f_t, g_t) \nabla_{\theta_m} d_{\cos}\Big(\mathcal{F}(s_{t+c}) - \mathcal{F}(s_t), g_t(\theta_m)\Big), \tag{9}$$

**WORKER of Generator**

$$\nabla_{\theta_w} \mathbb{E}_{s_{t-1} \sim G}\Big[\sum_{x_t} r_t^I \mathcal{W}(x_t | s_{t-1}; \theta_w)\Big]$$

$$= \mathbb{E}_{s_{t-1} \sim G, x_t \sim \mathcal{W}(x_t | s_{t-1})}[r_t^I \nabla_{\theta_w} \log \mathcal{W}(x_t | s_{t-1}; \theta_w)], \tag{10}$$

$$r_t^I = \frac{1}{c} \sum_{i=1}^{c} d_{\cos}\Big(\mathcal{F}(s_t) - \mathcal{F}(s_{t-i}), g_{t-i}\Big). \tag{11}$$

**Pseudo Code**

---

**Algorithm 1** Adversarial Training with Leaked Information

---

**Require:** Hierachical policy $G_{\theta_m, \theta_w}$; discriminator $D_\phi$; a sequence dataset $\mathcal{S} = \{X_{1:T}\}$
1: Initialize $G_{\theta_m, \theta_w}, D_\phi$ with random weights $\theta_m, \theta_w, \phi$.
2: Pre-train $D_\phi$ (i.e. the feature extractor $\mathcal{F}(\cdot; \phi_f)$ and the output layer $sigmoid(\phi_l, \cdot)$) using $\mathcal{S}$ as positive samples and output from $G_{\theta_m, \theta_w}$ as negative samples.
3: Pre-train $G_{\theta_m, \theta_w}$ using leaked information from $D_\phi$
4: Perform the two parts of pre-training interleavingly until convergence.
5: **repeat**
6:    **for** g-steps **do**
7:       Generate a sequence $Y_{1:T} = (y_1, \ldots, y_T) \sim G_\theta$
8:       **for** $t$ in $1:T$ **do**
9:          Store leaked information $f_t$ from $D_\phi$
10:         Get Q($f_t, g_t$) by Monte Carlo Search via Eq. (8)
11:         Get the computed direction $g_t$ from MANAGER
12:         Update WORKER parameters $\theta_w, \psi$, softmax via Eq. (10)
13:         Update MANAGER parameters $\theta_m$ via Eq. (9)
14:       **end for**
15:    **end for**
16:    **for** d-steps **do**
17:       Use current $G_{\theta_m, \theta_w}$ to generate negative examples and combine with given positive examples $\mathcal{S}$
18:       Train discriminator $D_\phi$ for $k$ epochs by Eq. (2)
19:    **end for**
20: **until** LeakGAN converges

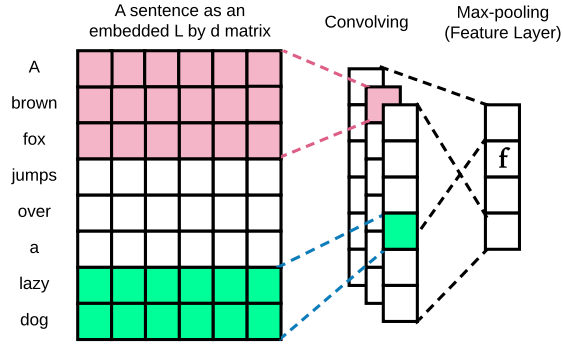

Figure 1: The feature extractor's architecture (without the highway and dropout layer)

## Experiment Settings

For synthetic data with length 20, the learning rate for MANAGER and WORKER is set to 0.001. The goal dimension size $k$ is set to 16. The embedding size of the LSTM-RNNs is set to 32. For the discriminative model, we set the hyperparameters of the CNN as Table 1

For synthetic data with length 40, the learning rate for MANAGER and WORKER is set to 0.0005. The goal dimension size $k$ is set to 16. The embedding size of the LSTM-RNNs is set to 32. For the discriminative model, we set the hyperparameters of the CNN as Table 1

Table 1: Convolutional layer structures.

| Sequence length | (window size, kernel numbers) |
| --- | --- |
| 20 | (1, 100),(2, 200),(3, 200),(4, 200),(5, 200) (6, 100),(7, 100),(8, 100),(9, 100),(10, 100) (15, 160),(20, 160) |
| 40 | (1, 100),(2, 200),(3, 200),(4, 200),(5, 200) (6, 100),(7, 100),(8, 100),(9, 100),(10, 100) (16, 160),(20, 160),(30, 160),(40,160) |

## Discussions

**The Necessity of the Hierarchical Architecture**  The hierarchical architecture in LeakGAN serves as the mechanism of incorporating leaked information from $D$ into $G$. However, in the body part, we haven't shown whether the explotation of hierachical architecture is a must. Actually, what we have to point out is, the explotation of hierarchical reinforcement learning is not a must, but a good choice in sequence decision scenarios.

We attempt to replace the hierarchical architecture by a fully connected layer. However, the model is so numerically sensitive that we cannot operate a stable training on it the original training settings. A possible reason is that, since the feature space of CNN changes rapidly during the training procedure, linear transformation without any normalization may not be able to incorporate the information contained in the feature vector leaked from $D$.

# Illustration of WORKER and MANAGER's Behaviors

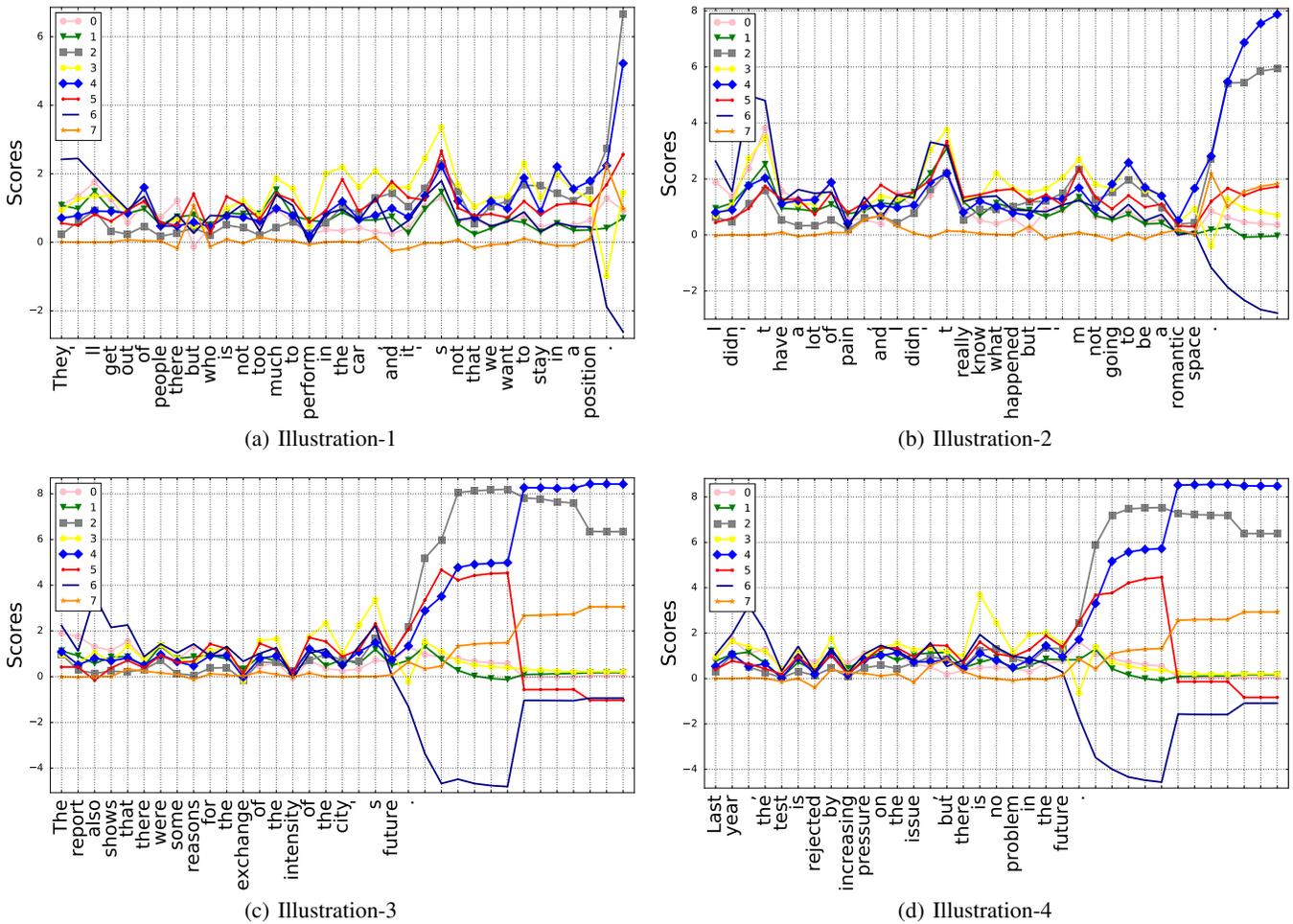

Figure 2: Illustration of WORKER and MANAGER's behaviors during a generation. (Dimension-wise Product of Worker and Manager)

Here we present more examples for illustrating the interaction of WORKER and MANAGER to support our claims in the main text as below. Each curve shows a subscore of the token of that time step. Each dimension of the score, i.e. each subscore measures a specific feature of the token in that context.

(i) The 5th dimension stands for current token's divergence from an entity token. If the 5th value is high, the token would most possibly be a structural token, such as a modal verb, an article or a preposition.

(ii) The 6th dimension suggests how long the suffix from current step will be. If a peak occurs in the curve, there must be some token that triggers a long suffix. A frequently occurring example is the formal subject.

(iii) Although hard to observe, we do find connections of the 7th dimension and the substructure of a sentence. For example, when the start or end of a sub-sentence occurs, there is an observable fluctuation in the 7th dimension. This indicates that the token is most likely to be a punctuation or a conjuction.

**Illustration of Feature Trace**

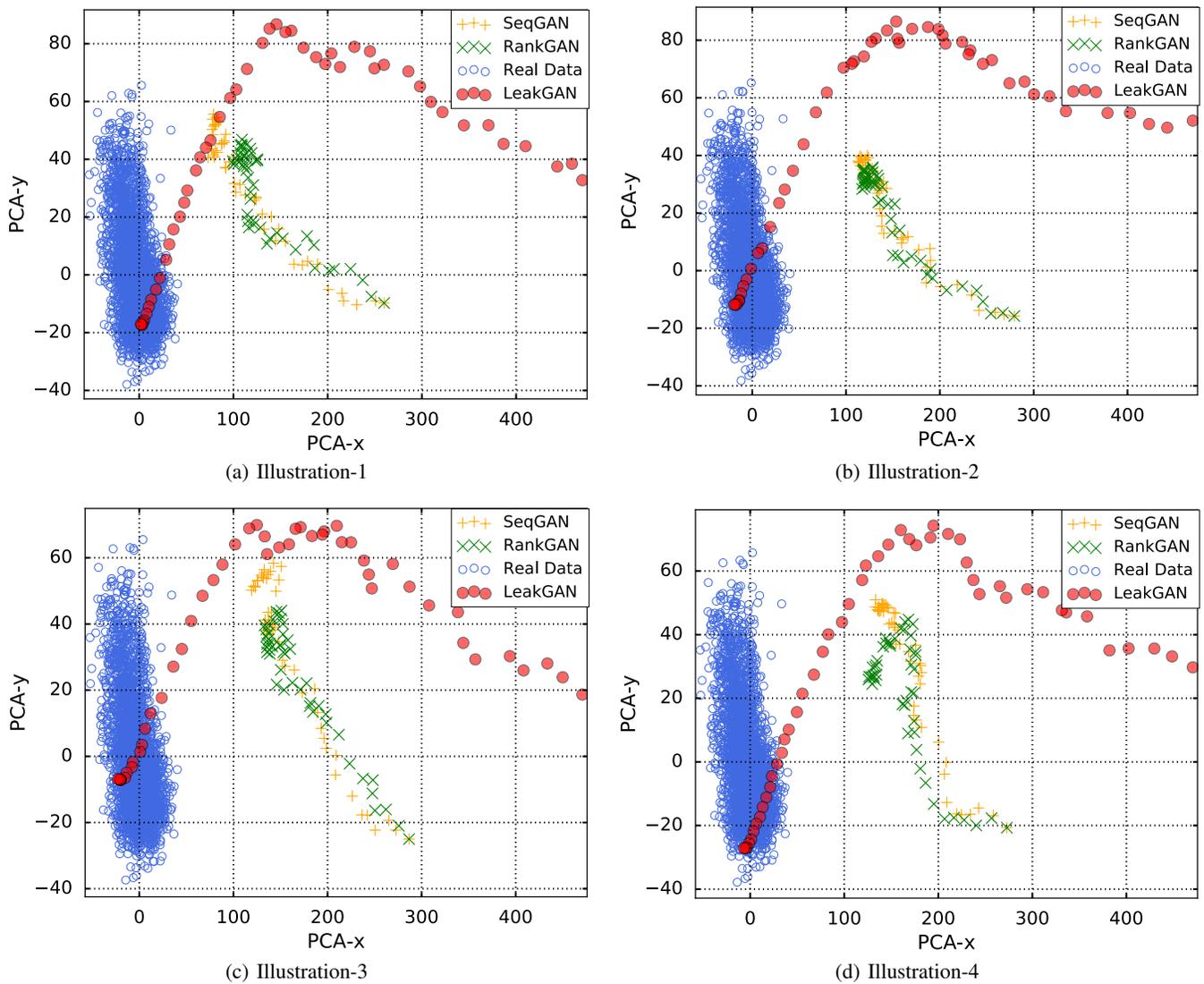

Figure 3: Feature traces (SeqGAN, RankGAN and LeakGAN) and features of real data (all compressed to 2-dim by PCA) on WMT News.

As we can see, during the generation process, in LeakGAN, the feature vector gradually approaches the real data feature vector region. However, previous models, i.e. SeqGAN and RankGAN, fail to match the features even when the generation is completed. This indicates that the proposed LeakGAN does finish its designed purpose of exploiting the leaked information from $D_\phi$ to better match the feature vector distributions of real data.

Table 2: Appendix 1 - COCO Examples in the Questionaire

| Sources | Example |
|---|---|
| Real data | A blue and white bathroom with butterfly themed wall tiles. |
| | The vanity contains two sinks with a towel for each. |
| | Several metal balls sit in the sand near a group of people. |
| | A surfer, a woman, and a child walk on the beach. |
| | A kitchen with a countertop that includes an Apple phone. |
| | A closeup of a red fire hydrant including the chains. |
| | People standing around many silver round balls on the ground. |
| | A person on a bicycle is riding in front of a car. |
| | A kitchen with a tile floor has cabinets with no doors, a dishwasher, a sink, and a refrigerator. |
| | The top of a kitchen cabinet covered with brass pots and pans. |
| | A woman is shaving her face while sitting on a wooden bench. |
| | A stuffed animal is laying on the bed by a window. |
| | A wooden toilet seat sits open in an empty bathroom. |
| | A person is taking a photo of a cat in a car. |
| | A phone lies on the counter in a modern kitchen. |
| | silver balls laying on the ground around a smaller red ball. |
| | A man riding a bicycle on a road carrying a surf board. |
| | A man using his bicycle to go down a street. |
| | A set table with silverware, glasses and a bottle of wine. |
| | A large kite in the shape of the bottom half of a woman. |
| LeakGAN | A woman holding an umbrella while standing against a sidewalk. |
| | A bathroom with a toilet and sink and mirror. |
| | A train rides along the tracks in a train yard. |
| | A man with a racket stands in front of a shop window. |
| | A red and white photo of a train station. |
| | The bathroom is clean and ready for us to use . |
| | A man is walking with his dog on the boardwalk by the beach. |
| | A man in a shirt and tie standing next to a woman. |
| | A couple of luggage cart filled with bags on a shelf. |
| | Large white and clean bathroom with white tile floors and white walls . |
| | A group of people fly kites in the sky on a clear day. |
| | A man wearing a suit and coat holds a tie through and wood pants. |
| | Two men are working on a laptop in a room . |
| | A man who is standing next to a brown and white horse. |
| | A street sign with a red stop sign on the street pole. |
| | A cat is laying on a keyboard and mouse in the air. |
| | A man with a rainbow - colored shirt and a black dog. |
| | A crowd of people standing around or standing on a sidewalk. |
| | A man is sitting on his desk holding an umbrella. |
| SeqGAN | A woman is riding a bike on the street next to a bus. |
| | A silver stove, the refrigerator, sitting in a kitchen. |
| | A guy doing tricks on a skateboard while a man is standing on a cellphone. |
| | A bunch of birds that are sitting in the sand. |
| | A bathroom with tiled walls and a shower on it. |
| | A couple of people are riding bikes down an asphalt road. |
| | An old photo of a man riding on a motorcycle with some people. |
| | A beautiful young girl in the bathroom has one has wine glasses and bottles above the counters. |
| | A person in a helmet standing next to a red street. |
| | An empty clean bathroom with a toilet and sink and tub. |
| | A kid in a black shirt and dog arms in a restaurant kitchen. |
| | A bathroom has a toilet, a sink and mirror. |
| | Two bicycles are parked outside inside a small brown field. |
| | The large rug is on the city under the city. |
| | A bathroom that is has a picture above and a sink. |
| | A small child jumping with glasses to a motor scooter. |
| | A white bathroom with a toilet, television and bathtub and a sink. |
| | A baby in a blue dress standing in front of a Frisbee. |
| | A cat and a woman standing by two computer preparing food. |
| | A pair of skis and pedestrians in a parking area near some different go. |
| | Two bikes in a parking lot with a dog that has a back on her. |

Table 3: Appendix 2 - News Examples in the Questionaire

| Sources | Example |
| --- | --- |
| Real data | Out of those who came last year, 69 per cent were men, 18 per cent were children and just 13 per cent were women. |
| | ' Sometimes I think about leaving sex work, but because I am alone living costs are really expensive,' she said. |
| | ' I was then stuck in the house for nearly two years only going out for short periods of time,' she said. |
| | He has not played for Tottenham's first team since and it is now nearly two years since he completed a full Premier League match for the club. |
| | This is a part of the population that is notorious for its lack of interest in actually showing up when the political process takes place. |
| | I was paid far too little to pick up a dead off of the ground and put it back in the box. |
| | Local media reported the group were not looking to hurt anybody, but they would not rule out violence if police tried to remove them. |
| | The 55 to 43 vote was largely split down party lines and fell short of the 60 votes needed for the bill to advance. |
| | We got to a bus station in the evening, but our connection didn't leave until the following morning. |
| | It's actually something that I had to add, because I was getting really frustrated losing to my hitting partner all the time. |
| | Taiwan's Defence Ministry said it was "aware of the information," and declined further immediate comment, Reuters reported. |
| | Her response to the international refugee crisis gave a million refugees hope that they may be able to begin a new life. |
| | I'm racing against a guy who I lost a medal to - but am I ever going to get that medal back ? |
| LeakGAN | A man has been arrested at age 28 , a resident in Seattle , which was widely reported in 2007 . |
| | I also think that ' s a good place for us , I ' m sure that this would be a good opportunity for me to get in touch . |
| | What is the biggest problem for Clinton is that Donald Trump will be in the race and he ' s unlikely to be the nominee . |
| | " We ' re going to do and we ' re going to put it out and get the ball ," he said . |
| | " I would be afraid to blame the girls to go back but I was just disappointed with the race," he said. |
| | " I'm not going to work together with a different role and we can win the game," he added. |
| | The couple's lives are still missing and they have been killed in the city's way to play against them, and because I came out there. |
| | For the last three years, we've got a lot of things that we need to do with this is based on the financial markets. |
| | Don't ask me, but I know, if I' ll be able to be out of Hillary Clinton, I think it's being made for the Congress. |
| | " I am proud to be able to move forward because we don't have to look at about," he said. |
| | That ' s why we ' re the most important people for the African American community and we ' ve made a good response . |
| | But the move will be only in a fight against them, as well as likely to prevent an agreement to remain in the EU. |
| | The American Medical Association said that the militants had been arrested in connection with the murder of the same incident. |
| | The two - year - old girl has been charged with a suspect who was in the vehicle to the police station. |
| | It is hard to buy on the Olympics, but we probably don't see a lot of it. |
| | " I'm not going to be very proud of the other countries," he said. |
| | He said the U. N. intelligence industry will not comment on the ground, which would be sensitive to the European Union. |
| | I take my work in the days, but I would have to go down on Wednesday night. |
| SeqGAN | You only certainly might not rush it down for those circumstances where we are when they were the heads, and when she's name. |
| | " I think you should really really leave for because we hadn't been busy, where it goes to one," he wrote. |
| | All the study knew was that they are, so they continue to provide support service and it doesn't exist. |
| | ' It can say become up with nothing sales have reached the charge for the other any evidence that been virtually well below the $ 800. |
| | Three times before the start of the season is much early on 2015 we are in the third training every year. |
| | That's the idea of strength that decision they said, we haven't already lost four or seven, or Liverpool's team. |
| | That is not the time for the cost of changing the system and it was pushing for $ 20 million. |
| | We had to take it a good day for a military, but nearly 6, 000 ] and prepare for them through. |
| | I actually didn't tell the background check the difference after my hour was to be recalled... and it was great. |
| | We are thinking about 40, 000 and jobs in what is wrong in the coming and you know. |
| | That is out how working you can't set out some pretty tight... or what I'm going through. |
| | " I wanted to be made you decided to have a crisis that way up and get some sort of weapon, not much to give birth to for an American room. |
| | She had been fined almost 200, 000 with couple of asylum seekers in Syria and Iraq. |
| | Perhaps not, in looking for, housing officials would help the frustration of Government, with an FBI shortly before 2020. |
| | Once we got to real show for the young man since I'm sure she went to love it just, whether to be late later last year. |
| | But, after a holiday period we might have to go on a total - out debate like that could have happened to us. |